\documentclass[journal]{IEEEtran}
\usepackage{amsmath,amsfonts}
\usepackage{algorithmic}
\usepackage{algorithm}
\usepackage{array}
\usepackage[caption=false,font=normalsize,labelfont=sf,textfont=sf]{subfig}
\usepackage{textcomp}
\usepackage{threeparttable}
\usepackage{stfloats}
\usepackage[inline]{enumitem}
\usepackage{acronym}
\usepackage{url}
\usepackage{verbatim}
\usepackage{graphicx}
\usepackage{cite}
\usepackage{xcolor}
\def\BibTeX{{\rm B\kern-.05em{\sc i\kern-.025em b}\kern-.08em
		T\kern-.1667em\lower.7ex\hbox{E}\kern-.125emX}}
\DeclareMathAlphabet{\altmathcal}{OMS}{cmsy}{m}{n}

\begin{document}
\title{Generative data augmentation for biliary tract detection on intraoperative images}

\author{Cristina Iacono$^{*}$, Mariarosaria Meola$^{*}$, Federica Conte, Laura Mecozzi, Umberto Bracale, Pietro Falco, Fanny Ficuciello
 \thanks{This research has been funded partially by the European Union – NextGenerationEU under the LeCoR-PROC project, PRIN 2022 PNRR, identification code Prot. P20229CSJB, CUP: E53D23014540001, partially by the European Union – NextGenerationEU under the TI-RED  project, PRIN 2022, identification code Prot. 20229N5YW8, CUP: E53D23012310006 and partially by the European Union - ERC-2023-SyG Endotheranostics (n. 101118626)} 
\thanks{Cristina Iacono is with the C.R.E.A.T.E. Consorzio di Ricerca per l'Energia, l'Automazione e le Tecnologie dell'Elettromagnetismo, Università degli Studi di Napoli Federico II, Napoli, Italy.} 
 \thanks{Mariarosaria Meola, Laura Mecozzi and Fanny Ficuciello are with the Department of Information Technology and Electrical Engineering, Università degli Studi di Napoli Federico II, Napoli, Italy}
 \thanks{Federica Conte and Pietro Falco are with the Department of Information Engineering, University of Padova, Padova, Italy.}
 \thanks{Umberto Bracale is with the General Surgical Unit, Department of Public Health, Università degli Studi di Napoli Federico II, Napoli, Italy}
 \thanks{$^{*}$ Cristina Iacono and Mariarosaria Meola are co-first authors.}
 \thanks{Manuscript received April 19, 2021; revised August 16, 2021.}
}
\maketitle


\acrodef{gan}[GAN]{Generative Adversarial Network}
\acrodef{ai}[AI]{Artificial Intelligence}
\acrodef{cnn}[CNN]{Convolutional Neural Network}
\acrodef{dcnn}[DCNN]{Deep Convolutional Neural Network}
\acrodef{icg}[ICG]{Indocyanine Green}
\acrodef{cv}[CV]{Computer Vision}
\acrodef{iou}[IoU]{Intersection over Union}
\acrodef{lc}[LC]{Laparoscopic Cholecystectomy}
\acrodef{ls}[LS]{Laparoscopic Surgery}
\acrodef{mata}[MATA]{Prisma Dataset}
\acrodef{mis}[MIS]{Minimally Invasive Surgery}
\acrodef{mirs}[MIRS]{Minimally Invasive Robotic Surgery}
\acrodef{rcnn}[R-CNN]{Region-based Convolutional Neural Network}
\acrodef{yolo}[YOLO]{You Only Look Once}
\acrodef{ccu}[CCU]{Camera Control Unit}
\acrodef{wl}[WL]{white-light}
\acrodef{bdi}[BDI]{Biliary duct injuries}
\acrodef{nir}[NIR]{near-infrared}
\acrodef{hsv}[HSV]{Hue Saturation Value}
\acrodef{rgb}[RGB]{Red Green Blue}
\acrodef{map}[mAP]{Mean Average Precision}
\acrodef{tp}[TP]{True Positive}
\acrodef{fp}[FP]{False Positive}
\acrodef{fn}[FN]{False Negative}
\acrodef{ap}[AP]{Average Precision}
\acrodef{dl}[DL]{Deep Learning}
\acrodef{dcnn}[DCNN]{Deep Convolutional Neural Network}

\begin{abstract}
Cholecystectomy is one of the most frequently performed procedures in gastrointestinal surgery, and the laparoscopic approach is the gold standard for symptomatic cholecystolithiasis and acute cholecystitis.
In addition to the advantages of a significantly faster recovery and better cosmetic results, the laparoscopic approach bears a higher risk of bile duct injury, which has a significant impact on quality of life and survival.
To avoid bile duct injury, it is essential to improve the intraoperative visualization of the bile duct.
This work aims to address this problem by leveraging a deep-learning approach for the localization of the biliary tract from white-light images acquired during the surgical procedures. To this end, the construction and annotation of an image database to train the Yolo detection algorithm has been employed. Besides classical data augmentation techniques, the paper proposes Generative Adversarial Network (GAN) for the generation of a synthetic portion of the training dataset. Experimental results have been discussed along with ethical considerations.
\end{abstract}


\section{Introduction}

\ac{lc}, gold standard technique for patients with gallstone disease, is one of the most commonly performed routine procedures in general surgery worldwide. \acp{bdi} are dangerous complications following \ac{lc} associated with significant postoperative effects related to morbidity, mortality and long-term quality of life~\cite{angelis2021e}~\cite{cirocchi2021}.
During \ac{lc}, a common cause of BDIs is the failure to clearly identify the anatomical landmarks within the hepatocystic triangle, particularly in distinguishing between the cystic artery and the biliary tract~\cite{akbari2009image}. 
To mitigate this risk, several measures have been implemented over the years~\cite{strasberg2002avoidance}. 
Among the others, the use of fluorescence imaging during intraoperative navigation~\cite{sincavage2024indocyanine} and the application of \ac{ai} and \ac{cv}  techniques for real-time surgical video analysis~\cite{guo2023}. Clinical experience with fluorescent tissue dye has been identified as a safe, non-toxic method for accurately visualizing tissue areas of anatomical interest where the dye has accumulated, thus enhancing operative speed, safety and patient outcomes~\cite{boni2015}.
In particular, \ac{icg}-enhanced fluorescence imaging, due to its good spatial and temporal resolution, has been introduced in laparoscopic surgery as a real-time method to evaluate the organ perfusion, providing detailed anatomical information during surgery~\cite{li2023detection}\cite{osborne2017}.

Since human bile contains albumin and other lipoprotein-binding \ac{icg}, the dye is rapidly excreted via the bile duct, allowing for the easy intraoperative recognition of biliary tract anatomy\cite{alander2012}. However, exclusive reliance on \ac{nir} imaging is impractical because only the \ac{icg}-enhanced area appears bright, leaving the surrounding tissue dark. Consequently, surgeons must frequently alternate between \ac{wl} and \ac{nir}. Integrating \ac{wl} to initially locate the target region could therefore simplify the surgical procedure, leading to improved outcomes and enhanced patient safety~\cite{kumar2016laparoscopic}\cite{osayi2015near}. Nevertheless, the major issue in using \ac{icg} is the risk of spreading to adjacent tissues during operation, limiting the technique’s sensitivity in biliary duct localization~\cite{endo2023}.
As previously mentioned, rapid advancements in \ac{cv} and \ac{ai}, particularly the widespread use of \ac{dl}, which employs multi-layered neural networks for complex task learning, have revolutionized the analysis of intraoperative video~\cite{ward2021}. Outstanding performances in medicine have been observed in areas such as surgical intervention planning, disease progression monitoring, and treatment optimization, as well as in the detection and diagnosis of cancer and polyps through CT and colonoscopy exams. In particular, \ac{ai}-assisted surgery has also been developed for \ac{lc}~\cite{madani2023}, enhancing surgical safety and quality, especially for novice surgeons~\cite{Wu2024}. However, the real-time surgical guidance from images and video is still more complex due to the variability in terms of noise, quality and objects within the field~\cite{diwan2023}\cite{guo2023}.
Object localization is a \ac{cv} problem that involves determining an object’s position inside an image or video and creating a bounding box of the object to indicate its presence and location~\cite{kumar2021optimized}. This process basically relies on estimating parameters from image features, using generative or discriminative methods. The first ones, such as particle filtering, mean shift and optical flow, detect objects by minimizing the difference between the modeled appearance of the identified target and potential candidates. In contrast, discriminative methods (e.g., support vector machines, decision forests, and \ac{cnn}s learn to distinguish targets from background by training a classifier on both. Due to their robustness and ability to effectively differentiate between target and background, discriminative approaches have become the leading techniques in object identification and tracking. 

The computational capabilities of GPUs allowed the transition to \ac{dl}-based detection, leading to significant breakthroughs in object detection, improving the overall performances~\cite{zou2023object}\cite{srivastava2021}. 
Object detection algorithms commonly used in medicine leverage \ac{cnn}s for their ability to automatically extract features from extensive training datasets, enabling accurate predictions. This has led to the development of various \ac{cnn}-based methods in recent years for identifying and tracking laparoscopic instruments and organs, especially during \ac{mis}~\cite{benavides2024}.

As reported in literature \cite{patel},  \ac{dcnn}-based detectors can be categorized as either "two-stage" or "one-stage". 
Two-stage detectors separate object localization (generating region proposals) from object classification, whilst one-stage detectors perform both detection and classification simultaneously using \acp{dcnn}. 


In addition, while the first ones generally offer superior \ac{ap}, single-stage detectors are faster, making them better suited for real-time applications. \\Given that the application area influences which algorithm is picked, a state-of-the-art, real-time, end-to-end object detection algorithm gained significant attention for its real-time performance in object detection tasks: \ac{yolo}~\cite{ragab2023}. Unlike \ac{rcnn} or its variants, \ac{yolo} looks at the whole image once to perform the detection. 
This regression-based object detector consists of a single \ac{cnn} that simultaneously predicts bounding boxes and their class probabilities from the image pixels. Furthermore, \ac{yolo} can encode contextual information by processing the entire image during training, leading to fewer background errors, compared to \ac{rcnn} or Fast \ac{rcnn}. In addition, it is extremely fast and suitable for detecting or localizing objects in real time. However, \ac{yolo} revealed some limitations, such as the need for strict constraints on the bounding box predictions and the difficulties in detecting small objects in an image. Over the years, \ac{yolo} has undergone numerous updates, each bringing forth a range of improvements and optimizations aimed at enhancing both the speed and accuracy of the model, preserving its efficacy~\cite{le2023}. Due to its characteristics, \ac{yolo} has proven highly effective for real-time object localization in healthcare, especially used as autonomous recognition method for
multiple instruments tips in surgical scenarios \cite{peng2022}. However, since it is based on a \ac{cnn}, a major drawback for medical object detection is its requirement for a large dataset of images for training to ensure optimal performance and avoid the risk of dataset bias, overfitting and inaccurate results. Nevertheless, the collection of these high-quality data can be both time-consuming and expensive~\cite{ragab2023}. 

To address such difficulties, data augmentation strategies have been proposed to improve robustness and generalization of the trained models by artificially inflating the training dataset size~\cite{padigela2023comparison}. 
Based on the taxonomy proposed in~\cite{shorten2019}, two main groups of methods can be distinguished: the transformation from original data and the generation of artificial ones. In the first case, in addition to erasing, elastic and pixel-level alterations, the affine transformations like cropping, flipping, or translation are the most popular types~\cite{goceri2023}. 

However, these traditional methods have inherent limitations, such as computation overhead and storage, and can only reshape existing data, generating new samples that are too similar to the original dataset, resulting in a slight improvement in diversity. 
Over the past few years, \acp{gan} have been established as a novel, promising, deep learning architecture for the synthesis of realistic-looking synthetic data. Their operational paradigm is analogous to a game, where a generator aims to create data identical to real data and a discriminator critiques it by distinguishing between authentic and synthetic images. This adversarial training dynamic compels the network to progressively enhance the realism and fidelity of the synthetic data it produces.

The medical research community has harnessed the potential of this approach to generate synthetic, realistic medical images across various modalities and applications ~\cite{chen2022}, aiming to expand original datasets, address class imbalances and enhance model robustness.
For instance, in fundus imaging, GANs have been used to synthesize realistic retinal (or neuronal) images. This was achieved by converting hidden, binary annotations of vascular or neuronal shapes into complete, visually convincing pictures \cite{Pinto2019}. Similarly, multi-parametric magnetic resonance images of abnormal brains (e.g. with tumors) have been generated from segmentation masks of brain anatomy and tumor \cite{shin2018}. The technique has also been applied to augment high-resolution whole breast images achieving high level of fine-grained details \cite{ren2019}. 
Furthermore, GANs have been employed to expand mid-sized thoracic X-ray datasets. Salehinejad et al. \cite{salehinejad2018} trained separate generators with random latent vectors to produce full X-ray images, specifically augmenting data for various lung diseases. 

However, one recurrent theme is how \acp{gan} can be effective in generating useful medical data, so that an evaluation of true utility and benefits~\cite{Skandarani2023}.
 
In this study, a deep-learning algorithm based on \ac{yolo}v11 for the detection and localization of the biliary tract in \ac{lc} has been implemented. A dataset consisting of labeled \ac{wl} and \ac{nir}-\ac{icg} videos acquired during standard laparoscopic surgical procedures has been collected, and the extracted frames were used for training. Different data augmentation techniques, based on both geometric/photometric transformations and \acp{gan} were applied to expand the original dataset. An analysis has been conducted to evaluate the performance of the models in terms of the anatomical landmark detection degree based on independent operators’ annotations.

\begin{figure}[htbp]
    \centering
    \begin{minipage}{0.49\linewidth}
        \centering
        \includegraphics[width=\linewidth]{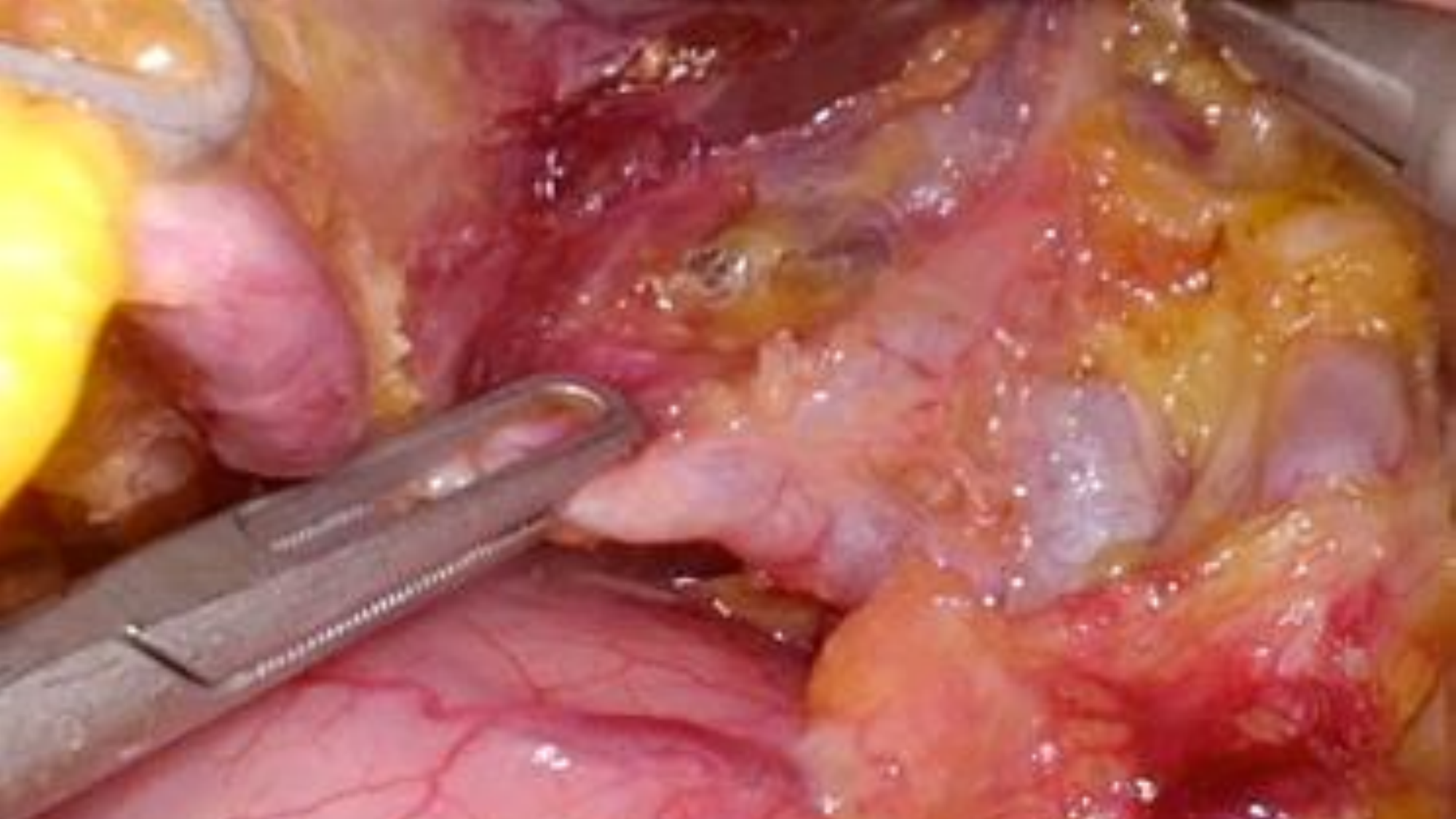}
        \label{fig:immagine1}
    \end{minipage}
    \hfill
    \begin{minipage}{0.49\linewidth}
        \centering
        \includegraphics[width=\linewidth]{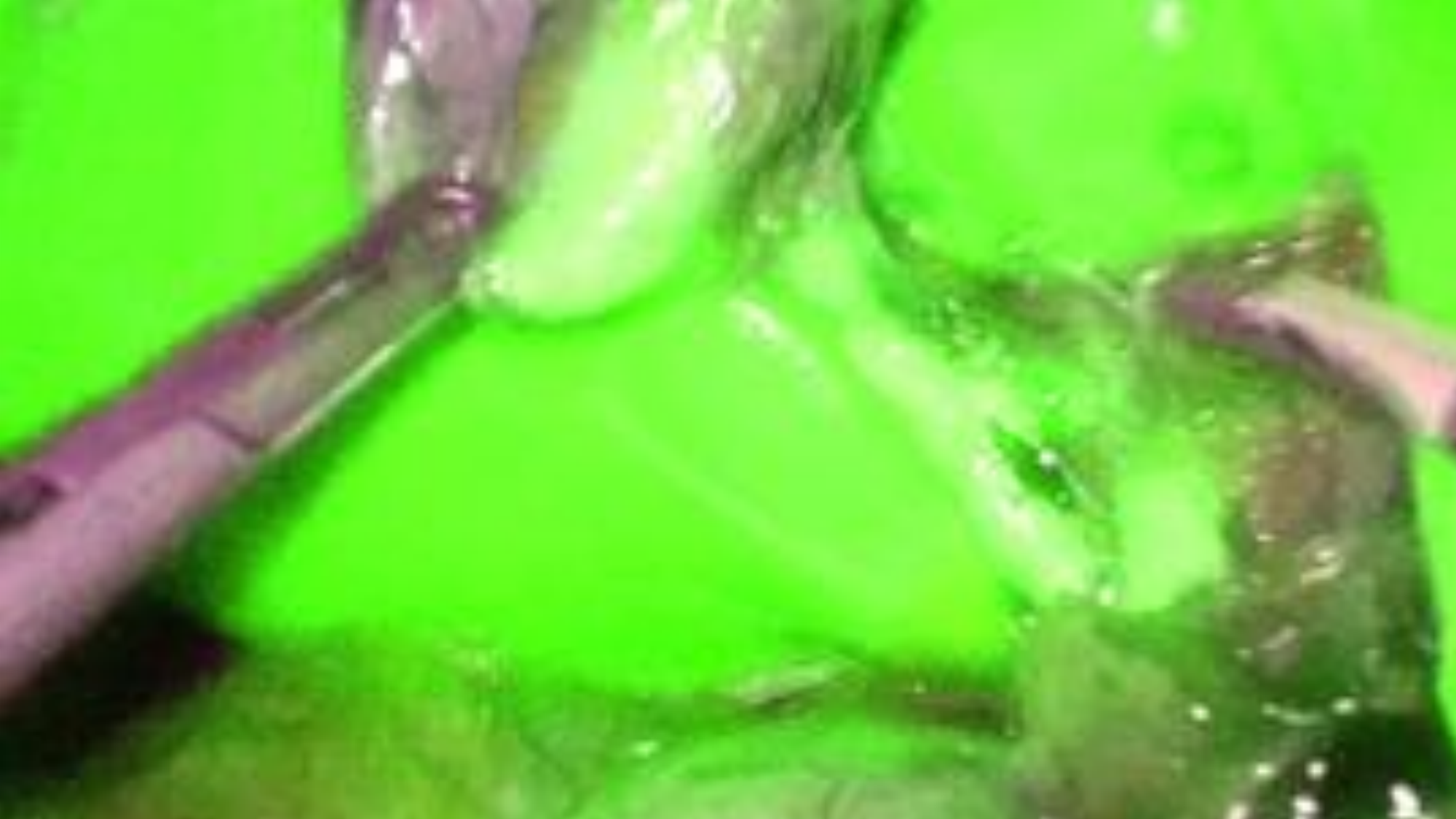}
        \label{fig:immagine2}
    \end{minipage}
    \caption{Representative intraoperative WL (left) and NIR-ICG (right) bile duct images.}
    \label{fig:confronto}
\end{figure}

\section{Materials and Methods}
\subsection{YOLOv11}  

To localize the biliary tract from the endoscopic images, the latest iteration from Ultralytics, namely \ac{yolo}v11, has been employed. Compared to previous versions, significant advances are introduced: improved architecture for enhanced feature extraction and optimized training pipelines for improved processing speed~\cite{yolo11_ultralytics}. 
Thus, this latest version is able to achieve higher precision, speed, and efficiency and it is designed to address various computer vision applications, including object detection, instance segmentation, and oriented object detection.

In this work, the \ac{yolo}v11 model has been trained for object detection, i.e. identifying both the class and location of objects in an image or video. The output consists of the detected object, enclosed within a bounding box, along with the class label and confidence scores for each detection. In particular, \ac{yolo}11s, the model pre-trained on the COCO Val2017 dataset~\cite{lin2014microsoft}, has been utilized, which achieved the following performance metrics:  
\begin{itemize}  
   \item \ac{map}val(50-95) of 47;  
   \item Inference speed: $2.5 \pm 0.0\,ms$.  
\end{itemize}  
 This pre-trained model served as the basis for further training. Subsequently, the pre-trained model was reused on the dataset of interest to exploit its pre-trained weights, following the fine-tuning approach.

\subsection{{Evaluation Metrics}}
To evaluate the performance of the object detection model the following metrics have been employed:
\begin{itemize}
    \item \acf{iou}: It measures the overlap between the bounding box predicted by the model and the ground-truth bounding box. It is a metric that ranges from 0 to 1, where a value of 0 indicates no overlap between the boxes, while a value of 1 indicates perfect overlap. During training, it is possible to set an \ac{iou} threshold to determine when a detection is considered accurate. A practical threshold value is 0.5, which can be adjusted depending on the training objective. It is computed as the ratio between the intersection area of the ground truth and predicted boxes and their union area~\cite{rosebrock2016intersection}. 
        \begin{equation}
            IoU = \frac{IntersectionArea}{UnionArea}
            \label{eq:iou}
        \end{equation}
    \item Precision: It evaluates the model's ability to avoid false positives, by providing a measure of the true positives among all those that were predicted as such. It is computed as the ratio between \ac{tp} and the sum of \ac{tp} and \ac{fp}.
    \begin{equation}
        Precision=\frac{TP}{TP+FP}
    \end{equation}
    \item Recall: it provides a measure of the true positives among all the actual objects of the class. This metric evaluates the model's ability to detect all relevant objects. It is computed as the ratio between i \ac{tp} and the sum of \ac{tp} and \ac{fn}.
    \begin{equation}
        Recall=\frac{TP}{TP+FN}
    \end{equation}
    \item \acf{map}: It integrates Precision and Recall into a single metric. It computes the \ac{ap} for each class, which is the average of precision values at various recall levels. The mAP is then obtained by averaging the \ac{ap} across all classes in the dataset. It provides an overall measure of how effectively the model detects and classifies objects across different classes. Higher mAP values indicate a better performance of the model.
        \begin{equation}
        mAP=\frac{1}{n}\sum_{i=1}^{N}{AP_i}
    \end{equation}
    In particular, the following versions have been considered:
       \begin{itemize}  
       \item \ac{map}50: It represents \ac{map} calculated with an \ac{iou} threshold of $0.50$. This metric evaluates the object's presence but allows loose localization.  
       \item \ac{map}50-95: It represents \ac{map} calculated over multiple IoU thresholds from $0.50$ to $0.95$ with a step size of $0.05$.
       It provides a more comprehensive assessment of the model’s performance, as it evaluates it across different levels of difficulty.  
   \end{itemize}

\end{itemize}

\section{{Dataset}}  

\begin{table}
\centering
\caption{DATASET COMPOSITION}
\begin{tabular}{|c||c|c|c|}
\hline

 & \textbf{\textit{White}}& \textbf{\textit{Green}} & \textbf{\textit{GAN}}  \\
\hline
Patient 1 & - & 231 & -  \\
\hline
Patient 2 &	- & 199	& - \\
\hline
Patient 3& - & 200 & -	\\
\hline
Patient 4& - & 204 & -	\\
\hline
Patient 5&	34	& - & 34	\\
\hline
Patient 6& 7 & 58 & 7	\\
\hline
Patient 7& 39 & - & 39	\\
\hline
 Patient 8& 15 &	- & 15	\\
\hline
 Patient 9&	74	& - & 74	\\
\hline
 Patient 10&	5 & - & 9 \\
\hline
 Patient 11 & 28	& -	& 57	\\
\hline

\end{tabular}
\label{tab:patients}
\end{table}

The dataset consists of videos collected from 11 patients who underwent \ac{lc} procedures between 2020 and 2021. 
The acquisition system consists of a \ac{ccu} connected to a camera head by an integral cable. The high-definition endoscopic camera system, sensitive in the visible and infrared spectrum, was used to acquire intraoperative images in \ac{wl} and \ac{nir}-\ac{icg} mode. The optical images were then transferred from the surgical site to the camera head by means of rigid and flexible scopes. The videos were acquired using a $25\,Hz$ frame rate with an average duration of $30\,s$. 

The complete dataset is highly heterogeneous: it comprises \ac{wl} and \ac{icg} videos and has been augmented using multiple techniques. Moreover, it is completely anonymous, as no patient's personal information has been stored. The videos collected from patients who presented complications that did not fall within the scope of this study were rejected. 

The resulting dataset consisted of 1329 640x640 resolution video frames extracted sampling 1 frame every 10, as shown in Table \ref{tab:patients}. 
The frames were manually annotated to delineate bounding boxes on the bile duct. For each image, in accordance with the \ac{yolo} guidelines, a text file has been created with the following information:
\begin{enumerate*}[label={(\roman*)}]
    \item \textit{class ID} of the object starting from 0;
    \item normalized \textit{X center} coordinate of the bounding box;
    \item normalized \textit{Y center} coordinate of the bounding box;
    \item normalized \textit{width} of the bounding box;
    \item normalized \textit{height} of the bounding box.
\end{enumerate*}

The \ac{wl} images dataset has been obtained from only seven patients (Patient 5--11), which sometimes may not be enough for training a robust and generalizable neural network, since a limited dataset may lead to overfitting and poor model performance. 
To address this problem, data augmentation methods have been applied to both WL and NIR frames, increasing the analyzability of the dataset. This technique consists of adding an artificial supply of data, which could not be present in the real data, to allow generalization and optimization of the models, reducing the overfitting and increasing the robustness of the system~\cite{yang2022image}.

\subsection{{Geometric and Photometric transformations}}
\label{geo&phototrasform}
Several transformations have been applied to increase the original size of the dataset.
\textit{Geometric transformations} alter the space and layout of initial images, namely:
\begin{itemize}
    \item \textit{Zoom (0.5)}: It randomly zooms in or out by up to 50\% of the image size to help the model become more robust in the face of changes in object size and scale;
    \item \textit{Rotation (0.4)}: The images are rotated randomly within a range of $\pm0.4\deg$° to introduce variations in orientation, thereby facilitating the model's learning of rotational variance;
    \item \textit{Translation (0.1)}: It translates images by up to $\pm10\%$ of their height or width, enabling the model to recognise variations in translation and simulate the displacement of the subject from the scene.
    \item \textit{Shear (0.1)}: The image is tilted by up to $\pm0.1\deg$ in order to ensure the system's robustness to changes in perspective, tilt and/or camera distortion;
    \item \textit{Horizontal Flip (0.5)}: The images are then subjected to a random horizontal flip with a probability of 50\%, thereby enhancing the variability and generalisation of the dataset.
\end{itemize}

These transformations are designed to introduce variations in the spatial organization and distribution of the images and to simulate different positions and distortions that the model might encounter when working in real-world scenarios.

\textit{Photometric Transformations} have also been applied to the dataset, altering the \ac{rgb} channels of the images, namely:
\begin{itemize}
    \item \textit{Tonal (0.015)}: this technique changes the color of the image in the \ac{hsv} scale up to $\pm{1.5}\%$, to handle white balance variations;
    \item \textit{Saturation (0.7)}: changes the saturation up to $70\%$, in order to manage images with more or less intense colors;
    \item \textit{Brightness (0.4)}: changes the brightness up to $40\%$ and is useful to make the model more robust to light variations.
\end{itemize}
    
Finally, image combination techniques were used, such as: 
\begin{itemize}
    \item \textit{Mixup (0.3)}: This combines two images with a compound label. It is active in 30\% of cases and helps generalisation;
    \item \textit{Mosaic (1.0)}: combines four images into one with a probability of $100\%$ to help the model learn to recognize objects in different sizes or variable positions. 
\end{itemize}

\subsection{GAN} \label{sec:gan}
To further enhance the original dataset, \ac{gan}-generated images have also been employed. 
In particular, \textit{HistaugGAN}~\cite{wagner2021structure}, a specialized \ac{gan}, designed for medical image augmentation, has been applied to create high-quality synthetic images by using a generator network against a discriminator one. The generator creates new images based on learned patterns, while the discriminator evaluates their authenticity. This iterative process is used to improve the quality of the synthetic data, making the model more efficient and robust.
By incorporating \textit{HistaugGAN}, the dataset can be expanded, and it is possible to introduce diverse and realistic examples, improving the model’s performance and generalization capabilities.

The method allows for the generation of images that preserve the structural integrity of the samples. Hence, it learns to simulate realistic variations in staining and imaging conditions, making the dataset more diverse and representative of real-world scenarios. 

Before training the \ac{gan}, the images undergo a preprocessing pipeline to ensure consistency and improve training stability. The raw images are first loaded and converted to the RGB color space, standardizing the input format. They are then resized to a fixed dimension while preserving the aspect ratio to prevent distortions. To facilitate faster convergence and enhance numerical stability, pixel values were normalized. The preprocessing procedure helps create a well-conditioned dataset, improving the quality of the generated images and the overall performance of the model. After processing through the model, the images are resized back to their original dimensions and saved for further use. These steps ensure a robust augmentation pipeline, enhancing dataset diversity and improving model generalization.

\begin{figure}[htbp]
    \centering
    \includegraphics[width=1\linewidth]{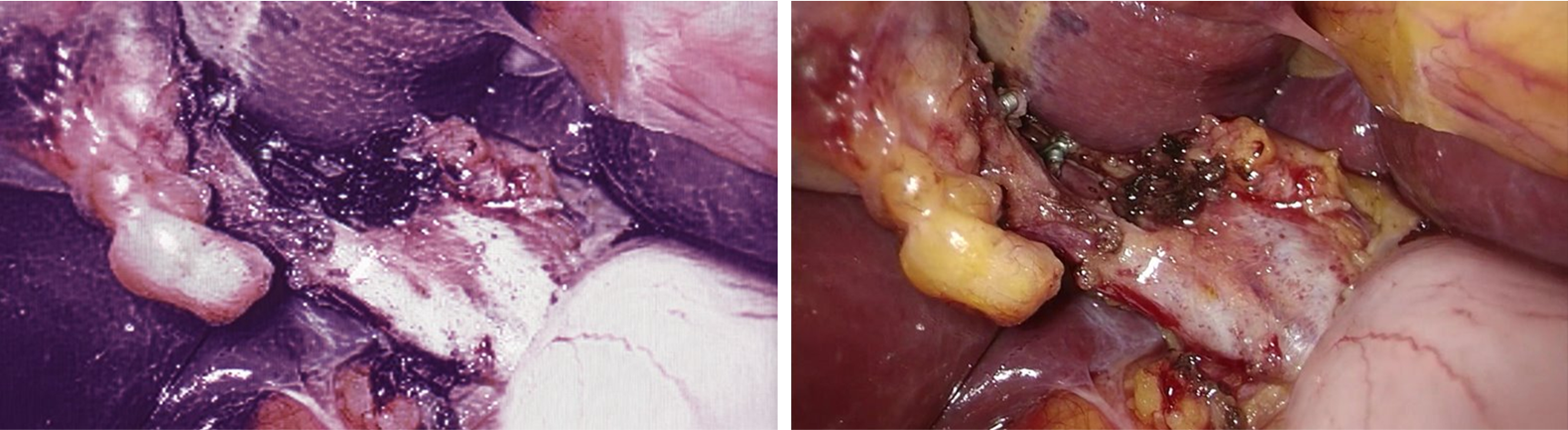}
    \caption{Example of GAN generated image from the WL original image.}
    \label{fig:GAN_example}
\end{figure}

\section{Experimental Settings and Results}
The experimental validation comprises a set of tests designed to analyze the impact of data augmentation on the localization algorithm's performance using the original dataset.
Based on the unsatisfactory results obtained during the preliminary testing phase  (the \ac{map}50-95 equalling to 0.003 and an \ac{iou} of 0 , indicating no detection) NIR-ICG images were excluded from subsequent analyses, and their findings are therefore not presented.
The \ac{yolo} algorithm was trained locally on a Dell Precision 3660 tower with 32 GB of RAM, a 13th-generation Intel Core i7-13700 processor, and 1 TB of storage.
Then, the \ac{yolo} algorithm has been trained with three different sets of images and the results are shown and discussed below.
Specifically, the following versions of the dataset were used:
\begin{itemize}
    \item \ac{wl} images;
    \item \ac{wl} + \ac{gan};
    \item Dataset of \(50\)\% \ac{wl} images + \ac{gan} images.
\end{itemize}
\begin{table}[h]
    \centering
    \caption{Dataset.}
    \renewcommand{\arraystretch}{1.2} 
    \begin{threeparttable} 
    \begin{tabular}{|c|c|}
    \hline
        \textbf{Dataset} & \textbf{Images} \\
    \hline
        WL & 202 \\
        WL + GAN & 367 \\
        WL50\tnote{*} + GAN & 265 \\ 
    \hline
    \end{tabular}
    \begin{tablenotes}
        \item[*] 50\% of original image set. 
    \end{tablenotes}
    \end{threeparttable}
    \label{tab:datasets}
\end{table}
Table \ref{tab:datasets} enumerates the total number of images used for each model, using a 90/10 split for training/validation and testing respectively. The test set is always composed of the images from Patient 6, which have been completely excluded from the training/validation stage, and a mix of images from the other patients up to 10\% of each dataset. Moreover, the test images are only \ac{wl} since the ultimate goal is to deploy the model for real-time use on patients, and therefore, it is not of interest to test the model on the generated images, which are only introduced and used to evaluate their impact on the predictions. 

\begin{figure*}
    \centering
    \includegraphics[width=1\linewidth]{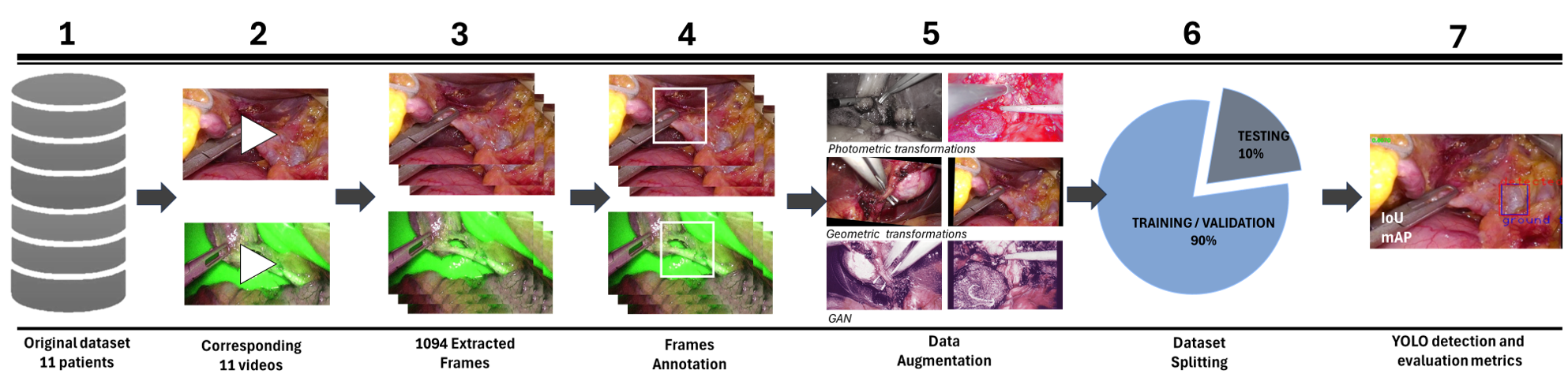}
    \caption{YOLO-based pipeline for biliary tract detection. The initial dataset comprised 1094 frames extracted from 11 intraoperative WL and NIR videos, corresponding to 11 subjects. This original dataset was expanded through various data augmentation strategies after annotation. The YOLO algorithm was trained and validated using a 90/10 split (training and validation/testing) and performance was evaluated using IoU and mAP. As explained within the manuscript, following step 6, the NIR dataset was excluded from further validation. }
    \label{fig:pipeline}
\end{figure*}

A complete training, with batch size set to 16, has been conducted both with and without the addition of manipulation operations as described in Section~\ref{geo&phototrasform}. To avoid overfitting and underfitting, Early Stopping, already implemented in the YOLO network, was used and set to \(100\). 
The proposed approach for the evaluation of data augmentation strategies using YOLO applied to intraoperative images is described in Figure\ref{fig:pipeline}. 

\subsection{{White Light Training Set}}
In this experiment, the model has been trained on the \ac{wl} images dataset, and these results were used as a benchmark for subsequent tests. In particular, 182 \ac{wl} images have been used for training and validation, while the test set is composed of 20 images comprising all the images from Patient 6  and a mix of frames from the other patients that were excluded from the training.

\begin{figure}
    \centering
    \includegraphics[width=1\linewidth]{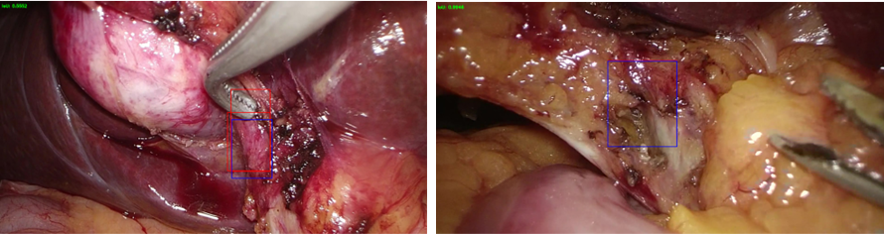}
    \caption{Worst (left) and best (right) results in terms of IoU for WL images trained with data augmentation. IoU = 0.5552 and IoU = 0.9946, respectively. }
    \label{fig:Fig4}
\end{figure}

The results suggest a clear performance distinction between models trained with and without data augmentation (see Figure \ref {fig:Fig4}). Table \ref{tab:wl} shows the computed metrics without data augmentation in the first three rows and with data augmentation in the last three rows.
Given the relatively small size of the dataset, even though early stopping was employed, the results suggest that the algorithm may reach overfitting.
A second model was trained to address this limitation and promote better generalization, exploiting data augmentation techniques to enrich the dataset. This model showed performance metrics notably decreased on the test set, which, nonetheless, appear more realistic. These results are expected and suggest that the model, when challenged with a wider variety of inputs, learns more generalizable patterns rather than memorizing the specifics of the training data. This is further supported by the inclusion of a portion of Patient 6, who was excluded entirely from training and validation, in the test set, offering a more reliable assessment of the model's robustness. 

\begin{table}
\caption{WL without/with data augmentation}
    \centering
\label{tab:wl}
\footnotesize
\begin{tabular}{|c|c|c|c|c|c|}
    \hline
    \textbf{Phases} & \textbf{Precision} & \textbf{Recall} & \textbf{mAP50} & \textbf{mAP50-95} & \textbf{IoU} \\
    \hline 
     Train & 0.952  & 0.698 & 0.833 & 0.427 & - \\
     Validation & 0.934  & 0.7 & 0.837 & 0.434 & - \\
     Test & 1 & 1  & 0.97 & 0.90 & 0.97 \\ 
     & & & & & \\
     Training & 0.64  & 0.319 & 0.31 & 0.168 & - \\
     Validation  & 0.631  & 0.319 & 0.31 & 0.168 & -\\
     Test & 0.78 & 0.80 & 0.68 & 0.55 & 0.80 \\
     \hline
\end{tabular}
\end{table}

\subsection{GAN augmented training set}
This experiment consists of training the network both with the \ac{wl} images and the \ac{gan} generated images as detailed in Section \ref{sec:gan} to overcome the limitations of the original dataset. A total of 330 images have been used for training/validation comprising 158 \ac{wl} images and 172 \ac{gan} images, while 37 \ac{wl} images have been selected for the test set as in the previous training. 

Firstly, the model was trained on the combined WL+GAN dataset without applying further standard augmentations. The results, presented in Table \ref{tab:wl-gan} (rows 1-3), show that the model achieved a test \ac{map}50-95 of 0.82. This result, compared with the results in Table \ref{tab:wl}, demonstrates that enriching the dataset with new, plausible image instances is a more effective strategy for generalization than simply transforming existing images.

\begin{figure}
    \centering
    \includegraphics[width=1\linewidth]{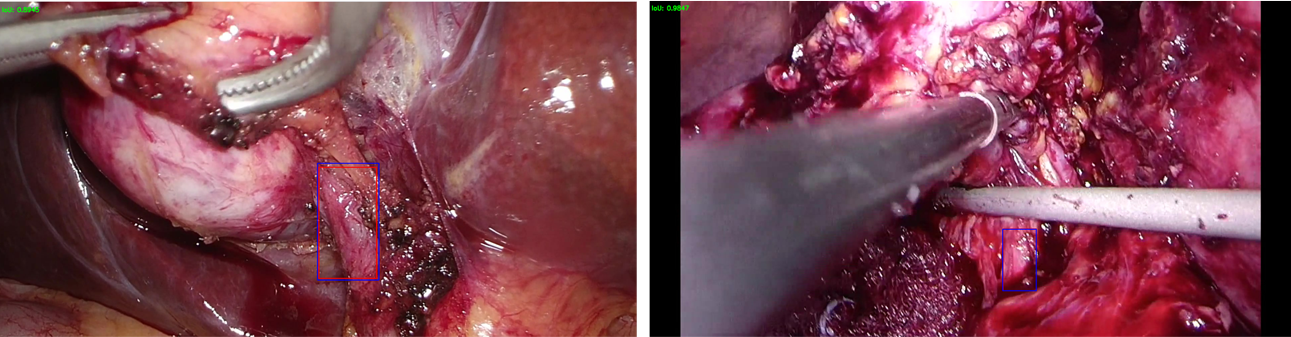}
    \caption{Worst (left) and best (right) results in terms of IoU for WL+GAN images trained with data augmentation. IoU = 0.8945 and IoU = 0.9847, respectively.}
    \label{fig:Fig5}
\end{figure}

Moreover, the results show that the combination of both strategies is the most successful approach. This model produced the best overall results among the generalizable models, as seen in Table \ref{tab:wl-gan} (rows 4-6), achieving a test \ac{map}50 of 0.96 and a \ac{map}50-95 of 0.88.  (see Figure \ref{fig:Fig5}).

The model's computational efficiency supports real-time applications, achieving an average inference time for biliary duct localization of $ 0.0755\,s$ and $ 0.071\,s$, respectively. 

\begin{table}[]
\caption{WL + GAN without/with data augmentation}
\label{tab:wl-gan}
    \centering
\begin{tabular}{|c|c|c|c|c|c|}
    \hline
    \textbf{Phases} & \textbf{Precision} & \textbf{Recall} & \textbf{mAP50} & \textbf{mAP50-95} & \textbf{IoU} \\
    \hline 
     Train & 0.947  & 0.297 & 0.411 & 0.221 & - \\
     Validation & 0.948 & 0.302 & 0.427 & 0.238 & - \\
     Test & 1 & 1 & 0.92 & 0.82 & 0.924 \\ 
     & & & & & \\
     Train & 0.781 & 0.343 & 0.412 & 0.237 & - \\
     Validation & 0.808 & 0.343 & 0.419 & 0.241 & - \\
     Test & 1 & 1 & 0.96 & 0.88 & 0.96 \\ 
     \hline
\end{tabular}
\end{table}

\subsection{GAN augmented training set with reduced WL set}
The last experiment has been designed to test the effects on the model of the ratio of real to synthetic data. To this end, the model has been trained using a reduced real-data portion of images, specifically, $50\%$ of the \ac{wl} images (66 WL images and 172 GAN).
Table \ref{tab:wl50-gan} shows the results obtained.

\begin{figure}
    \centering
    \includegraphics[width=1\linewidth]{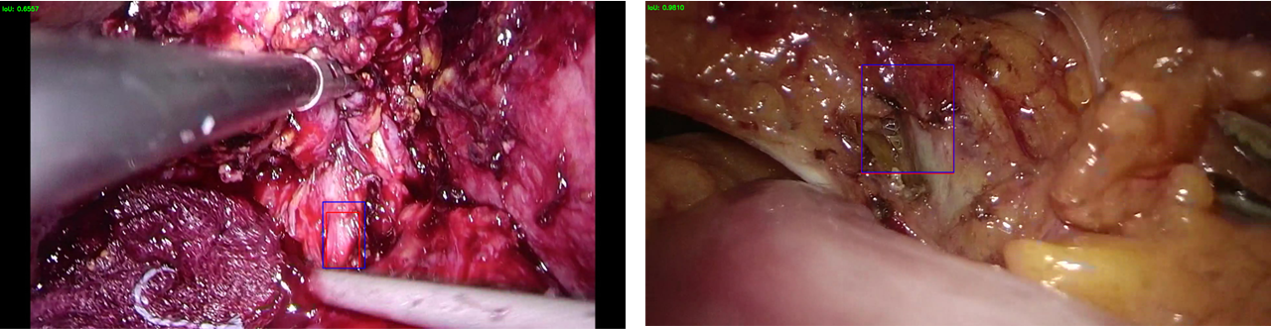}
    \caption{Worst (left) and best (right) results in terms of IoU for WL50+GAN images trained with data augmentation. IoU = 0.6557 and IoU = 0.9810, respectively.}
    \label{fig:Fig6}
\end{figure}

The model with data augmentation (rows 4-6 of Table \ref{tab:wl50-gan}) not only achieved very good performance but also surpassed the previous best model, scoring a \ac{map}50-95 of 0.89.  (see Figure \ref{fig:Fig6})
This result suggests that a linear increase in the amount of real data does not necessarily translate to an improvement in performance when synthetic data is utilized. Instead, there appears to be an optimal ratio of real-to-synthetic data that maximizes the model's generalization capability. 
By using a smaller amount of real data, the model is forced to rely more heavily on the diversity offered by the \ac{gan} images, preventing the model from becoming overly biased towards specific features present in the larger real dataset.

From a computational standpoint, the model achieves real-time performance, with an average inference time of approximately $0.071\,s$.

\begin{table}[]
\caption{WL50 + GAN without and with data augmentation}
    \centering
\begin{tabular}{|c|c|c|c|c|c|}
    \hline
    \textbf{Phases} & \textbf{Precision} & \textbf{Recall} & \textbf{mAP50} & \textbf{mAP50-95} & \textbf{IoU} \\
    \hline 
     Train & 0.665  & 0.234 & 0.252 & 0.172 & - \\
     Validation & 0.671 &0.235 & 0.309 & 0.211  & - \\
     Test & 1 & 1 & 0.96 & 0.89 & 0.96 \\ 
     & & & & & \\
     Train & 0.648  & 0.244 & 0.284 & 0.201 & - \\
     Validation & 0.648  & 0.244 & 0.284 & 0.201 & - \\
     Test & 1 & 1 & 0.96 & 0.89 & 0.958 \\ 
     \hline
\end{tabular}
\label{tab:wl50-gan}
\end{table}

\section{Discussion}  

Unlike usual approaches relying on NIR imaging, this research demonstrates YOLO's ability to detect bile ducts in WL images acquired during MIS, critically exploiting a GAN-enhanced dataset within the training step. \\
The expansion of the training dataset to 300 images through the integration of synthetic images generated by a \ac{gan} marks an important step toward improving the robustness and generalization of the \ac{yolo}v11 model in this medical imaging context.

In particular, the WL50 + GAN + Aug model emerged as the optimal configuration, outperforming all other setups. This result strongly suggests that for any specific problem, a "sweet spot" in the real-to-synthetic data ratio may exist, and the strategy of simply using "all available data" might not always be the most effective. For future work, the systematic exploration of this ratio presents a crucial avenue of research for the optimization of deep learning models. 

The calculated average inference time, slightly inferior to an endoscopic camera acquisition frequency,  demonstrates the potential to be employed to localize the biliary tract in real-time scenarios.

It is also essential to thoroughly discuss the ethical implications of using \ac{ai}, extending beyond detection to include dataset generation.

The progress of technology has amplified the human potential for knowledge. Thanks to \ac{ai}, decision-making processes, which typically involve people, are enhanced and transformed. However, this also brings new risks and ethical considerations. One only needs to think of issues strictly related to personal data privacy, cybersecurity, the quality of information, its integrity, and the necessity of addressing these topics with the awareness of not infringing upon fundamental human rights.

One of the main risks associated with \ac{ai} is given by the possible bias introduced during the training phase, leading to incorrect or distorted decisions. Biases can result in discrimination based on ethnicity, specific physical traits, or gender, implying different medical treatment for these categories. This risk is closely related to the data used for training the network, and it is especially true for small and dishomogeneous datasets. Data augmentation techniques may not be sufficient for neutralizing this possibility, since the generated data is still based on the original dataset.

Another limit of the use of \ac{ai} models is given by their transparency and interpretability. To mitigate risks, it is necessary to adopt responsible training practices that include data verification and, if necessary, data cleaning to correct biases. This requirement is specifically crucial for the use of \ac{ai}-generated data. In the case of point, the generated images may lack important features or may be overly simplified/smoothed. This can result in insufficiently nuanced details and diversity inherent in authentic datasets. Oversight of the data by multidisciplinary experts can help ensure that systems are developed in the most fair and responsible manner.

Finally, data security and privacy violations may occur if the stored dataset contains personal data, which can lead to privacy issues for the individuals involved \cite{eticAI}.

Therefore, specific regulations must be followed to ensure the use of \ac{ai} technologies with respect to fundamental rights. To this end, strict obligations are imposed on medical software and e-health solutions, defining acceptable and unacceptable risk levels \cite{aiact}.

\section{Conclusions}
This study proposed a method for the detection of the biliary tract from laparoscopic images. \ac{yolo}v11 has been trained on a manually labeled dataset from \ac{lc} procedures. \acp{gan}-generated images have been used to enhance the training dataset in order to improve the detection performance. 
The results demonstrate the efficacy of the synthetic generated data for improving the object detection in the proposed data-scarce setting. Moreover, the tests unveil the critical factor of an optimal balance between real and synthetic data. Further work should focus on investigating the factors that affect detection performance, including the balance between real and synthetic data used in the training stage.


\bibliographystyle{myIEEEtran}
\bibliography{IEEEabrv,bib}

\end{document}